# PARKINSON'S DISEASE MOTOR SYMPTOMS IN MACHINE LEARNING: A REVIEW

Claas Ahlrichs and Michael Lawo

Mathematics and Computer Science, University of Bremen,
PO Box 330 440, 28334 ,Bremen, Germany

## ABSTRACT

*This paper reviews related work and state-of-the-art publications for recognizing motor symptoms of Parkinson's Disease (PD). It presents research efforts that were undertaken to inform on how well traditional machine learning algorithms can handle this task. In particular, four PD related motor symptoms are highlighted (i.e. tremor, bradykinesia, freezing of gait and dyskinesia) and their details summarized. Thus the primary objective of this research is to provide a literary foundation for development and improvement of algorithms for detecting PD related motor symptoms.*

## 1. INTRODUCTION

This research focuses on algorithms for detecting Parkinson's disease (PD) related symptoms in time series data. PD is a disorder of the central nervous system resulting in a loss of motor function, increased slowness and rigidity. Artificial intelligence (AI)-based techniques can be utilized to detect symptoms such as tremor or bradykinesia while focusing on minimizing false negatives (i.e. failing to recognize a symptom) and false positives (i.e. detection of a symptom where none is apparent). Those affected by PD bear a great burden and have to cope with a rather reduced quality of life. In the authors' eyes, this is an even more pressing issue when considering leading role of Germany. In 2004, Germany inhabited the largest number of people with Parkinson's within Europe [3].

Even though it can manifest itself at any age, PD is among other diseases (e.g. Alzheimer's, dementia, chronic bronchitis) usually attributed to elderly subgroups of the population. Considering demographic changes of the last decades, the number of cases and burden of PD is expected to increase [24, p. 36]. The World Health Organisation (WHO) estimates that around 5.2 million people were suffering from PD worldwide in 2004 [40]. Depending on the estimating organization, Europe inhabited 1.2 [24] - 2.0 [40] million of them in the same year.

PD is typically characterized as a chronic, progressive, neurodegenerative disorder [4], [26], [58], [20], [27]. The cardinal symptoms are bradykinesia, rigidity, tremor and postural instability [26], [58], [27], [20], [23], [60], [3], [32]. Among many other symptoms, these symptoms result from a dopamine deficiency in the substantia nigra. A part of the brain that is located within the basal ganglia circuit (see Figure 1). Dopamine is a neurotransmitter involved in movement control [69]. Usually by the time of diagnosis, a great number of dopamine-producing neurons have already diminished [58]. Current treatments aim at slowing the progression of the disease, focus on symptomatic relief and attempt to lift the enormous burden of PD. However, a cure is yet to be found.

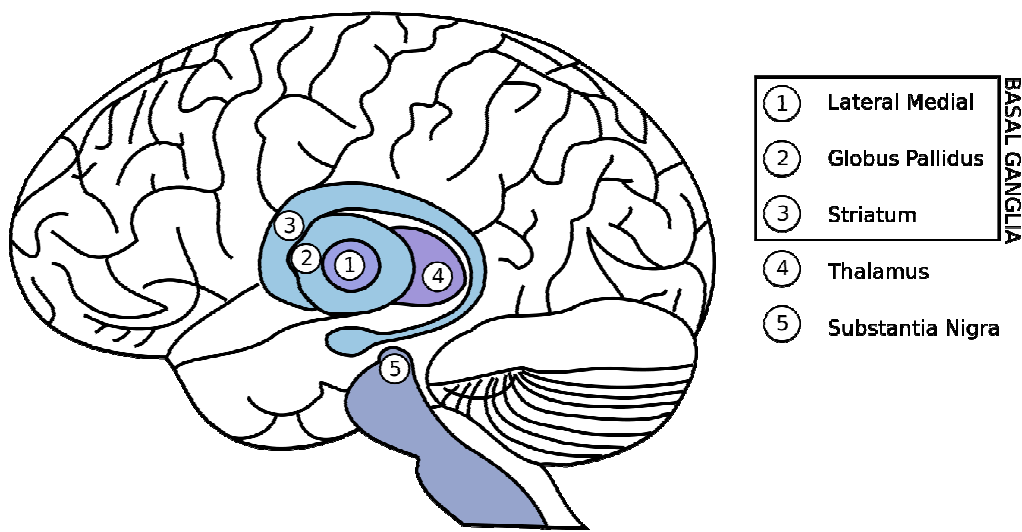

Figure 1. Illustrates structures of the brain related to the basal ganglia circuit and substantia nigra.

Latter is located in the upper end of the brain stem. The image is based on a figure which has been retrieved from Wikimedia Commons and belongs to the public domain.

PD is a great burden, not just for people suffering from the disease but also for those being indirectly affected (i.e. relatives and caretakers). In an advanced stage of the disease and without proper treatment, patients are no longer capable of taking care of themselves. In the Global Burden of Disease study, the WHO rated PD to be on the same disability level as: amputated arm, congestive heart failure, deafness, drug dependence and tuberculosis [40, p. 33].

A large number of symptoms have been shown by people with Parkinson's [27], [26]. The most visible and easily noticeable symptoms are related to motor functions. However, quality of life is affected by an even greater number of motor and non motor symptoms (e.g. depression, sleep disorder, cognitive / neurobehavioral abnormalities, autonomic and gastrointestinal dysfunction) [27], [26], [3]. As the disease progresses, patient's symptoms change and fluctuate (i.e. some symptoms simply disappear, while others (re-) appear), creating a unique symptomatic history for each individual patient. Unfortunately, in an advanced stage of Parkinson's further (drug-induced) symptoms may become apparent. Dyskinesia is one of these symptoms and results from a lengthy pharmacological treatment (i.e. several years). It manifests itself as an involuntary movement of entire body parts (e.g. rhythmical moving of upper body).

Tremor at rest (also known as rest tremor or resting tremor) is only present when muscles are at rest and dissolves during sleep as well as with action (i.e. voluntary movement of affected extremity) [34]. It manifests itself as an involuntary, unilateral (one-sided) shaking of an extremity (e.g. hand, foot, etc.). The shaking generally occurs at a frequency between 4-6 Hz [27].

Bradykinesia refers to slowness of movement [27], [34]. It usually appears in very early stages of the disease[58] and it is characteristic for basal ganglia disorders [27]. Depending on the severity, movements may not only be slowed (bradykinesia), but also diminished (hypokinesia) or completely abrogated (akinesia).

Freezing of gait (FOG) (also known as freezing or motor blocks) is a form of akinesia which presents itself as an inability to initiate or continue movement [27], [58]. Motor blocks are a common symptom, experienced by people with Parkinson's (although it does not occur uniformly) and can affect various extremities (e.g. arms and legs) as well as the face [27]. After onset of the symptom, it typically lasts for several seconds and disappears afterward. It is a common cause of falls [27], [58].

Publications reveal a great number of techniques for automatic detection of PD motor symptoms which employ various AI-based methods such as neural networks (NNs) [30], [15], [5], [9], [55], [21], [16], [14], hidden markov models (HMMs) [54] and support vector machines (SVMs) [9], [50], [14]. Depending on the symptom and utilized sensors, various features are calculated (e.g. entropy [9], [50], [14], [43], spectral or fractal features [64], [61], [66], [5], [57], [44], [47], [29], [10], [11]). Over time, sensor signals are analyzed and compared or set in relationship to known samples of each symptom in order to recognize them. No matter whether these AI methods are continuous or window-based, all of them can be viewed as either data mining techniques and / or time series (analysis) algorithms. Much literature presents algorithms for detecting a single symptom (e.g. [63], [66], [54], [5], [7], [45], [39], [21], [16], [14]). Considering the heterogeneous nature of symptom profiles in PD patients, this is not sufficient. Few publications focus on detecting of multiple motor symptoms (e.g. [57], [15], [55], [50]), but even those rarely consider enough symptoms for use in real-world scenarios. In reality, patients are likely to experience multiple symptoms, thus increasing the chance of false negatives and false positives.

To summarize, the objective of this research is to provide a literary foundation for development and improvement of algorithms for detecting PD related motor symptoms.

## 2. IDENTIFYING PARKINSON'S DISEASE AND ITS SYMPTOMS

Much research has been published with a focus on biological, chemical and genetic aspects of PD. Over the last two decades, an increasing number of publications originate from fields like computer science or AI, focusing on signifying motor symptoms in people with Parkinson's. Some of which are dedicated to detecting single motor symptoms [63], [66], [54], [5], [7], [45], [39], [21], [16], [14] and others on detection of multiple symptoms [57], [15], [55], [50]. These publications reveal a great number of techniques for automatically indicating the presence of PD motor symptoms (e.g. NNs [30], [15], [5], [9], [55], [21], [16], [14], HMMs [54] and SVMs [9], [50], [14]). Depending on symptom and utilized sensors, various features have been proposed and applied in literature (e.g. entropy [9], [50], [14], [43], spectral or fractal features [64], [61], [66], [5], [57], [44], [47], [29], [10], [11]) are known to be used in this context. In the course of this section a strong focus will be on the most common symptoms that are experienced by PD patients. Preference is given to those publications that do not focus on a single symptom (as opposed to multiple symptoms) or use synthetic datasets (as opposed to data recorded from sensors on the subject's body) but rather use unconstrained and unscripted activities of daily living (ADL).

It should be kept in mind that there are many other publications with a focus on PD symptom indication of their severity, but do not make use of body-mounted sensors or otherwise do not resemble closely the previously elaborated criteria. Despite this reservation, a few selected publications that do not fit these criteria are presented nonetheless.

## 2.1. TREMOR AT REST

In an early study by Salarian et al. [56], they were able to achieve a specificity of 98% and sensitivity of 76.6% on a dataset with ten patients and ten control subjects. In total, close to twenty hours worth of data were captured by the authors. Two tri-axial gyroscopes (i.e. one on each wrist) were used to record data while participants performed a set of scripted everyday activities. Spectral analysis was used to filter interesting regions within the specific frequency range specific to resting tremor(i.e. 3.5Hz-7.5Hz). In a later study [57] based on the same dataset, the data stream was divided into chunks with a length of three seconds to which the Burg method [13] was applied. Additionally a meta-analysis was introduced to remove isolated segments that were classified to exhibit tremor or tremor-like behavior (e.g. a single segment with tremor surrounded by none-tremor segments). This increased the sensitivity to 99.5% but lowered the specificity to 94.2%.

In [72], inertial sensor data (i.e. acceleration and angular velocity) were gathered from six patients and seven control subjects. Zwartjes et al. had captured approximately 1.5-2 hours worth of data while participants were performing a set of scripted activities in laboratory conditions. A multi-staged algorithm is utilized to indicate regions of tremor. At first some preprocessing is applied to the raw data, which is then used to classify the subject's activity and / or posture. This pre-classification is used to highlight regions of interest where tremor is more noticeable (e.g. arms are hanging still while standing). If activity or posture were usable for detection of tremor (and its severity) then those portions of the data stream are divided into segments of three seconds length with a two-thirds overlap. For each segment, the Fourier Transform is used to identify tremor specific frequencies and thus tremor episodes as well. In the algorithm's last stage, a meta-analysis removes isolated segments of tremor (very similar to the process that was utilized by Salarian et al. in [57]). Zwartjes et al. achieved an accuracy of about 84.7%. However, in comparison to studies by Salarian et al. [56],[57], the recorded activities were less constrained.

Rigas et al. [54] achieved an accuracy of 87% in detecting tremor in an accelerometer based dataset with twenty-three participants (i.e. ten patients and thirteen control subjects). All participants performed daily activities in laboratory conditions. The data stream is divided into three second windows with 50% overlap. Having applied standard filtering and analysis techniques (i.e. finite response (FIR) filters, Fast Fourier Transform, etc.) an HMM is utilized to detect tremor episodes. This is different from most algorithms for tremor indication. More common approaches rely on spectral features alone [64], [61], [66], [5], [57], [44], [47], [29], [10], [11] while other classify based on NNs [30], [15], [5], [9], [55], [21], [16], [14] or SVMs [9], [50], [14]. Rigas et al. state that HMMs are suitable for tremor indication because "tremor presents time-dependency" [54]. They consider HMMs as a time sensitivity extension of the naive Bayes classifier.

Cole et al. [15] were able to detect tremor with a sensitivity of 93% and specificity of 95% in unconstrained and unscripted activities. The dataset contained about 48 hours worth of acceleration and electromyogram (EMG) measurements from twelve participants (i.e. eight patients and four control subjects). Here a dynamic neural network (DNN) is used in combination with a set of FIR filters to detect tremor. It is stated by the authors that DNNs [67] were utilized because they are more capable of learning and classifying time-dependent classes (e.g. tremor) when compared to regular and static neural networks. Cole et al. divide the data stream into segments of two seconds length for feature extraction. The features are simply passed to the DNN, where artificial neurons do their work. However, the neurons' outputs are not simply forwarded to the next layer of neurons. Instead, each neuron has an FIR filter attached to it which transforms the output before it is passed to subsequent neurons. Their results are mainly dependent on the choice of training data. Here a handcrafted representative subset of data was chosen.

A dataset with nineteen patients and four control subjects was used by Roy et al. [55] to signify tremor. They achieved a sensitivity of 91.2% and a specificity of 93.4% in EMG and acceleration data. The participants were performing unscripted and unconstrained activities in a home-like environment for several hours. Here the data stream is also divided into two second windows and a combination of DNNs with FIR filters is fed with various features that were extracted from the two second segments.

Niazmand et al. [46] collected data from accelerometers integrated into a pullover. Ten patients and two healthy control subjects performed standardized PD motor tasks. An average sensitivity of 80% in indicating postural tremor and resting tremor and a specificity of 98.5% was achieved by the authors. Their algorithm first determines the relative acceleration among the sensors and then determines the movement frequency. This is done because sensors are not fixed on the patient's body but rather in a garment which position can change depending on executed movements. The raw data is simply filtered, normalized and a noise removal method is applied. For determining the movement frequency, a combination of thresholds and peak counting is utilized.

## 2.2. BRADYKINESIA

In [14], Cancela et al. present a motor symptom monitoring and management system. Their work originates from a European research project called PERFORM (Personal Health Systems for Monitoring and Point-of-Care Diagnostics-Personalized Monitoring). Here a set of classification algorithms (e.g. SVM, k-nearest neighbors (KNN), NN, decision tree (DT), etc.) was evaluated. The highest accuracy of 86% was achieved by the SVM. The corresponding dataset consists of acceleration data from twenty patients performing a set of ADL (within the limits of a scripted protocol). A standard analysis procedure is used by Cancela et al. At first a Butterworth filter is applied to raw sensor data then the data stream is epoched in five second segments with a 50% overlap. A set of features (i.e. sample entropy, root mean square, cross correlation, etc.) is calculated for each segment and passed to the classification algorithms. Here, the algorithms classify presence and severity of bradykinesia. Interestingly, the severity is not derived from standard motor tasks but instead from ADL.

Cancela was also involved in a publication by Pastorino et al. [49]. Here a slightly modified version of Cancela's algorithm is utilized (as in [14]). A dataset from twenty-four patients performing unconstrained and unscripted activities at their home was gathered for a week. Twice a day, a clinician came to visit the patient and performed a short protocolized session which was later used to test the previously developed algorithm. Pastorino et al. show that an additional meta-analysis can improve classification results. Instead of using the generated outputs from the SVM directly, they can be further filtered / smoothed to ignore impossible and unrealistic scenarios. Using a patient independent algorithm an accuracy of 68.3% ± 8.9% was achieved and a 74.4% ± 14.9% accuracy was achieved with the additional meta-analysis. They indicate that a patient specific training of the algorithms would likely lead to improved results.

Salarian et al. were not only involved in detecting tremor in time series data, they were also using gyroscopes on the wrists to indicate the presence of bradykinesia. In [56], ten patients and ten healthy control subjects participated in the collection of twenty hours worth of data. All participants were performing scripted ADL. Salarian et al. showed that the features rotation of hand ($R_H$) and mobility of hand ($M_H$) correlate well with the clinician's ground truth (r=-0.84 and r=-0.83 for $M_H$ and $R_H$ respectively and p<0.00001). Several years later, Salarian et al. were able to reproduce their results in [57]. However in latter publication, window sizes of five minutes and above were used. Even though their work does not produce results in real-time, it does give hope that not many sensors are required for a decent accuracy in bradykinesia detection.

The authors Zwartjes et al. [72] were able to identify bradykinesia related parameters that correlate well with the patient's unified Parkinson's Disease rating scale (UPDRS) scores. Here a dataset based on accelerometers and gyroscopes from six patients and seven healthy control subjects was analyzed. All subjects performed a mixture of standardized motor tasks and ADLs in a random predefined order. An activity / posture classifier is used to identify a set of elementary activities (i.e. walking, standing up) and postures (i.e. standing and sitting). For upper extremities, an average arm acceleration is calculated while various gait-related features (i.e. step length, step velocity, etc.) are determined from a tri-axial gyroscope and a tri-axial accelerometer that are placed on a foot. These features provide the basis for bradykinesia (slowness of movement) and hypokinesia (poverty of movement) quantification. The authors' results indicate that a significant correlation is present in almost all bradykinesia-related parameters while "none of the hypokinesia-related parameters were significantly correlated" [72].

## 2.3. AKINESIA / FREEZING OF GAIT

In [21], Djurić-Jovičić et al. employed a neural network and a simple thresholding technique to classify walking patterns in PD patients. A set of six inertial measurement units, each containing a tri-axial accelerometer and a tri-axial gyroscope, were attached to the subjects' legs (i.e. thigh and shin) as well as their feet. The kinematics of four patients (as they were following a predetermined path) were gathered, annotated and used to train a neural network. In total, about 30 minutes of data were collected. The path itself included several (potential) hurdles which have been designed to provoke FOG (e.g. start hesitation, destination hesitation, narrow path or turn hesitation, etc.). A combination of heuristically determined thresholds and a NN were utilized to differentiate between "normal" (i.e. standing and regular steps) and pathological (i.e. festination, akinesia, shuffling and small steps) walking patterns. The authors of [21] achieved an error rate as high as 16% due to the choice of thresholds (i.e. thresholds were independent of patients, etc.). On the contrary, the algorithm was working in real-time (i.e. about 0.5 seconds delay).

A similar technique was developed by Cole et al. in [16]. However, instead of a regular (static) neural network a dynamic neural network [67] was utilized. Their indicator algorithm showed a sensitivity of 82.9% and specificity of 97.3% in a dataset containing unconstrained and unscripted activities. Ten patients and two healthy control subjects contributed and helped to gather about two hours worth of data from several accelerometers (i.e. forearm, thigh and shin) and an EMG sensor (i.e. shin). The authors employed a multi-staged algorithm [16]. In the first stage a simple linear classifier determines whether the subject is in an upright position (i.e. standing and not sitting or lying). If this is the case then a DNN determines episodes of akinesia. The idea was to identify periods in the data stream were episodes of akinesia are more likely to be apparent (both visually and in data stream). In contrast to a static NN, Cole et al. [16] have decided to use a DNN because they are able to better capture time-varying weights that are present in FOG episodes.

An accelerometer based smart garment called MiMed-Pants [48] has been used by Niazmand et al. [47] to extract and analyze gait-related features. In this case, the measurement device has been successfully integrated in an item that is "suitable for daily use" [47]. The pair of pants can be washed like a regular textile. A sensitivity of 88.3% and specificity of 85.3% has been achieved with this setup. Five accelerometers (i.e. each shin, each thigh and belly button) provided kinematics on six patients while they were performing standard activities [71] (i.e. walking course, including narrow spaces, gait initialization and reaching destination, etc.). In total about one hour worth of data was collected by Niazmand et al. No use of advanced artificial intelligence methods was made in [47], but instead a linear classification was applied to features that were extracted from the sensors. The algorithms provided feedback with a delay of about two seconds.

In 2009, Bächlin et al. published their work on a wearable and context-aware system for real-time detection of FOG events [11]. The system provides acoustic feedback within a two second window. A set of accelerometers and gyroscopes was utilized (e.g. on shank, thigh and waist). Over eight hours worth of data were gathered from ten patients performing ADL, as well as walking in a straight line and a random walk. Bächlin et al. claim to have built the first context-aware and wearable system to assist PD patients in detecting FOG events. An overall sensitivity of 73.1% and 81.6% specificity were achieved in [11]. The results were mainly due to different walking styles (that were used by the subjects in their dataset), choice of features and use of patient independent thresholding. They state that a personalized training and choice of threshold might have produced better results.

In [62], Stamatakis et al. were able to show differences in walking patterns between a PD patient and a healthy control subject. The authors identified a set of features that may be used for differentiating between PD patients and healthy control subjects as well as for detecting FOG events and their duration. Even though no results in terms of accuracy or significance were presented in [62], their presented features may prove to be beneficial.

## 2.4. DYSKINESIA

Keijsers et al. [30] were able to achieve an accuracy of 96.8% in detecting dyskinesia. Thirteen participants were enrolled in their study. Each subject contributed about 2.5 hours of acceleration data while they were performing a set of scripted activities (approximately 35) in a controlled environment. In total six tri-axial acceleration sensors were attached to the subject's body (i.e. one on each thigh, one on each shoulder, one on trunk and one on wrist) during their recording session. The algorithm, used by Keijsers et al., classified fifteen minute segments using a regular neural network. The output of the NN indicated the presence (or absence) of dyskinesia within the segment. Several segment sizes were empirically evaluated (e.g. fifteen and one minute segments). The best accuracy was achieved with the fifteen minutes segments (i.e. 96.8%). However when using one minute segments the accuracy drops to about 80% on the same dataset.

In contrast, Tsipouras et al. [65] were able to achieve similar results but on smaller segments. While Keijsers et al. [30] used fifteen minute segments, here two second intervals with 75% overlap are utilized. Tsipouras et al. state to have achieved a 93.7% accuracy using their dataset. This contains inertial sensor data (i.e. acceleration and angular rate) of four patients and six control subjects. All participants were performing a set of scripted activities. In total two gyroscopes (i.e. one on trunk and one on waist) and six accelerometers (i.e. one next to each gyroscope, one on each arm and one on each leg) were used during recording sessions. Here five classification methods were evaluated (i.e. naive Bayes, KNN, fuzzy lattice reasoning, DTs and random forests (RFs)). The RFs performed best with 93.7%. C4.5 is close behind with about 93.5% while all other remaining classification algorithms achieved an accuracy around 85%.

Cole et al. [15] used a DNN to better capture the time-based variables. The authors were able to achieve a 91% sensitivity and a 93% specificity in detecting dyskinesia. Here a similar procedure to their work in detecting tremor [15] and bradykinesia [16] was utilized. Their dataset contained several hours' worth of acceleration data and EMG measurements from eight patients as well as four control subjects. Participants were performing unscripted and unconstrained activities during their recording session. A set of features is extracted from a two second sliding window and fed to the DNN. Additionally, outputs of each artificial neuron (i.e. node within the neural network) are filtered using a five point FIR filter.

Similarly Roy et al. [55] combine DNNs with a rule-based reasoning method. They were able to achieve a sensitivity of 90% and specificity of 93.4% in a dataset containing acceleration data and EMG measurements from nineteen patients and four control subjects. One hybrid sensor (containing a tri-axial accelerometer and an EMG sensor) was located on each arm and leg. All participants were performing unconstrained and unscripted activities in a home-like environment. In total about 30 hours worth of data were gathered and used by Roy et al. [55]. Again the extracted features originate from two second segments. Their algorithm uses those features to feed two DNNs (i.e. one for mobility states and one for motor states). These DNNs provide preliminary results on patient's mobility state (i.e. sitting walking, standing, etc.) and motor symptoms. They are used in combination with a framework called IPUS (Integrated Processing and Understanding of Signals), which in turn activates different DNNs to maximize symptom recognition rates (e.g. based on the fact that the subject is walking, sitting, etc.).

Patel et al. [51] utilized clustering techniques (and expectation maximization (EM)) to distinguish various levels of severity in PD patients while they were performing standardized motor tasks. A similar approach was used by Sherrill et al. [59]. Here six patients provided acceleration data to which clustering was applied in order to detect dyskinesia.

## 2.5 DYSARTHIA AND DYSPHAGIA

In [53], Revett et al. employ a rough sets approach for distinguishing healthy subjects and people with PD based on vocal data. Their dataset is based on thirty-one participants (i.e. twenty-three patients and eight healthy controls) performing a phonation task. Little et al. [36] originally constructed this dataset and donated it to the University of California Irvine (UCI) Machine Learning Repository [1]. On average each participant performed six phonations of the vowel [a]. Thus resulting in close to 200 samples of which a set of twenty-three features (including spectral features, shimmer, jitter, presence of PD, etc.) has been documented. Revett et al. report to have achieved a 100% accuracy when using all available features in their rough sets approach. This holds if the classification category is binary (i.e. healthy subject or PD patient). In this case several hundred rules are generated to identify a data sample's category. When trying to reduce the number of rules, the accuracy drops but stays well above 90% with about 100 rules. However, the authors did not attempt to perform classification based on the patient's duration of the disease, severity or UPDRS scores.

In a publication by Bakar et al. [8], they present a speech-based assessment tool for identifying PD. Here the same dataset (as originally constructed by Little et al. [36]) has been utilized. The authors performed several tests in which they compared testing accuracy, training accuracy, average mean square error (MSE) as well as average number of iterations of two training / learning algorithms for NNs (Levenberg-Marquardt (LM) and Scaled Conjugate Gradient (SCG)). Their results indicate that LM outperforms the SCG algorithm. Generally speaking, the LM-approach resulted in better testing and training accuracy as well as a lower MSE while SCG performed better in terms of "number of iterations". The best result was achieved with a 97.9% accuracy in training and 93.0% accuracy in testing.

Asgari and Shafran [6] utilized a similar set of features for classifying speech and phonation data. They performed a prediction of UPDRS motor scores based on these features. In more than one hundred recording sessions, twenty-one control subjects and sixty-one patients were asked to perform three tasks: sustained phonation (i.e. phonation of vowel [a]), diadochokinetic test (i.e. repetition of syllables [pa], [ta] and [ka]) and a reading task. The actual recordings were done with a dedicated device that was designed to be an at-home testing tool. The dataset itself was analyzed with 100 frames per second using Hamming windows (each of 25ms length). A rather large set of features is extracted from each frame and were generated for both voiced and unvoiced segments. About 15K features were generated based on their recordings of phonation and speech data (including pitch, frequency, harmony, etc.). A SVM was employed to translate these features into UPDRS motor scores. Depending on the used features (or subset of features) a mean absolute error between 6.1 and 5.7 (UPDRS) points was achieved.

In [41], Mekyska, Rektorova and Smekal evaluate a set of features for automatic analysis of speech disorders in PD. The authors provide an extensive summary on various speech-related parameters and highlight a few common problems in automatic speech analysis. Their dataset is composed of forty-two male control subjects and twelve male PD patients. Each participant was asked to pronounce all vowels (i.e. [a], [e], [i], [o], [u]) once in a natural speed and once slowly. The inter-intra class distance ratio method (IICDR) and minimum redundancy maximum relevance (MRMR) method were used by Mekyska et al. in order to sort out the top 20 parameters for each method (after having started with 510 parameters in total). In a second step, the Jarque-Bera test was utilized to see which features show a normal probability distribution. Those with a normal distribution were used in a multi-factor analysis of variance (ANOVA). Their results show up to three features (i.e. "mean B-$F_1$" for $p \leq 5\%$ as well as "mean $F_0$" and "mean NHR" for $p \leq 10\%$) that can be used to distinguish healthy control subjects from those afflicted with PD. Mekyska et al. add that there are also several parameters which do not show a coherent tendency in published literature. The authors point out several papers in which a particular feature has been shown to be significant, non-significant and indifferent in terms of separating patients from healthy subjects. However, they also comment that these conflicting publications usually used rather small datasets. Thus they are more prone to random variations and clusters.

Xiuming et al. [70] describe a diagnostic approach to PD based on principle component analysis (PCA) and Sugeno integral. The authors employ a dataset by Little et al. [36] from UCI Machine Learning Repository [1]. Thus data of thirty-one participants (i.e. eight patients and twenty-three healthy control subjects) was analyzed. Xiuming et al. show five principle components which account for 86.5% of the information within the signal. In order to propose a diagnosis, the Sugeno measure and Sugeno integral are then determined for the top five most relevant features. The authors report a classification accuracy of 81.0%.

## 2.6 OTHER

In 2000, Hamilton et al. [25] published their work on outcome prediction in pallidotomy in PD patients. It was their goal to build a reliable tool which estimates an operation's outcome based on intra-operational recordings of neural activity. A standard NN was employed and trained with a set of features (e.g. signal power, entropy or fractal dimensions). This system could provide supplementary data and aid surgeons in minimizing risks (e.g. blindness, difficulties in speaking or swallowing, etc.) and maximizing effectiveness. Their results indicate that all evaluated NN performed similarly in terms of overall outcome prediction. Hamilton and colleagues found that their NNs handled exceptional cases well.

In terms of diagnosis Kupryjanow et al. [35] came up with an alternative measurement technique for determining UPDRS sub-scores related to motor tests (i.e. finger tapping and rapid alternating movement of hands). Instead of relying on arguably subjective assessments from neurologists, they present a device called Virtual-Touchpad (VTP). Here a webcam is used to capture movements of hands and translate them into machine readable features. In comparison to other methods, this approach does not require equipment to be attached to the patient (or mounted on the patient). A SVM recognizes hand gestures and / or postures. The succession of those postures is used to extract the mentioned features and determine UPDRS scores. The authors did not describe a user study in [35].

In [18], Cunningham et al. presented their work on a computerized assessment tool. The work is intended to identify movement difficulties found in people with PD and similar movement disorders. Here the participants' ability to point and click on targets on the computer screen is compared among those afflicted with PD and healthy control subjects. A benefit of this approach is that patients are not expected to wear "unusual" or specialized hardware. However, on the other hand, it requires the patient to sit in front of a computer and cannot be mobile (as it would considerably alter their ability to point and click). Their results show a difference in control subjects and PD patients. The control group was generally more accurate (i.e. clicked closer to the target's center and made less accidental clicks) and faster (i.e. required less time to click once the target has been reached). This holds for both computer literates and computer illiterates. Those being computer illiterates and suffering from PD showed a higher variance in terms of accuracy of clicking the target center. In [19], Cunningham et al. present their tool's abilities in indicating akinesia, bradykinesia, dyskinesia, rigidity and tremor.

In a preceding publication by Cunningham and colleagues [17], another study has been performed with ten PD patients. Here participants were asked to use a home-based assessment tool twice a day (i.e. once in ON state and once in OFF state) for a period of four days. It is their goal to differentiate between a participant's ON state and OFF state based on their test performance. As in other studies by Cunningham et al. [18], the subjects clicked on targets while their speed, time, distance and location of click were recorded. Regarding the time, a statistical significance was found when comparing performances in ON and OFF states ($p = 0.017$). The authors also note that a few subjects showed an increased variance in ON-OFF state which indicates that they might not have been in a clearly defined ON-OFF state at the time of testing. Nonetheless their results appear to be promising and larger follow-up studies are to be seen.

Wang et al. [68] developed a new method for quantitative evaluation of symptoms in people with PD. Their proposed method is based on free spiral drawings with a digitizing table. Spiral drawings itself have been used a number of times to quantify motor dysfunctions (in particular [38], [37], [42] were highlighted by Wang et al.). However, these methods were usually employing some sort of guidance or template. In contrast, Wang et al. utilized free spiral drawings. A group of ten participants enrolled in their study at a hospital in Japan (Kaizuka). All of them were asked to draw a spiral which was then used to extract several features (e.g. number of turns, mean of radius, maximum radius, etc.). Their results indicate that healthy control subjects can be clearly separated from the remaining eight patients regarding the *number of extreme points in radius curve*. Most PD subjects were not able to "rapidly enlarge the circle as spiral" [68]. The authors remark that the features *mean value of radius* and *slope of radius curve* demonstrated stiffness and could be also used to distinguish between healthy and pathological subjects.

Pradhan et al. [52] considered a similar methodology, but instead of drawing spirals thirty PD patients were tracking waves by applying force to sensors. Here two force sensors (i.e. one for index finger and one for thumb) were "squeezed" in order to track a wave (i.e. simple sine wave and complex wave with multiple frequency components) with and without mental distraction. Their goal was to provide an assessment tool for clinical progression of PD patients. Pradhan and colleagues state that similar studies have been performed but "which may not be effective in documenting subtle changes in motor control" [52]. When comparing their task of wave tracking (involving precision control), other studies were usually employed to quantify surgical results or treatment progression. Three features were considered: *spectral density*, *root mean square (RMS) error* and *lag*. Although some of the features correlated significantly with UPDRS scores, there have been no significant improvements in prediction. Nonetheless, the authors suggest that their test may add an extra objective measure that other tests fail to capture.

In a publication by Brewer et al. [12], a similar approach has been used to predict UPDRS scores. Here twenty-six participants (all PD patients) were exhibiting pressure on force and torque sensors while they were performing wave tracking tasks. The authors used the same parameters to summarize the participants ability to properly track waves (i.e. spectral density, RMS error and lag). These features were evaluated in terms of their ability to predict UPDRS scores. The authors present four approaches: PCA, least squares linear regression, lasso regression and ridge regression. Their results indicate that ridge regression works best with an absolute error of 3.5 UPDRS points. This is followed by lasso regression (i.e. 4.5 UPDRS points) and PCA (i.e. 7 UPDRS points).

Similarly, Kondraske et al. [31] utilize ordinary computer hardware for specialized PD tests. The authors present an initial evaluation of three objective, self-administered and web-based tests (i.e. alternating movement quality, simple visual-based response speed and upper extremity neuromotor channel capacity). Each test has an equivalent version in the real-world based on a testing device called "BEP 1". Twenty-one subjects (i.e. eight healthy controls and thirteen PD patients) enrolled in their evaluation where both lab-based and web-based tests were performed. The results indicate an encouraging well correlation by lab-based and web-based "rapid alternating movement" and "`neuromotor channel capacity" tests. The correlation for the "simple visual" test did not show expected results. The authors envision a three-tiered approach that first involves digital, web-based tests then lab-based tests and finally screening by an expert. As suggested by its nature, web-based tests are easily accessible to a broad population. They provide objective measurements within an uncontrolled environment and may provide an initial assessment on whether any signs of PD are apparent. The second tier can then be used for a complementary assessment in a controllable environment. Afterwards a proper clinical screening can be performed by a neurologist if previous results suggested parkinsonian behavior.

An automatic evaluation approach for early detection of PD is presented by Jobbágy et al. [28]. The authors propose and evaluate a set of tests that were specifically designed to highlight features of PD symptoms. They employ a motion tracking system, called precision motion analysis system (PRIMAS), for recording movements patterns. The system uses a combination of infrared (IR) light, passive markers (i.e. small, lightweight reflective disks mounted on body) and cameras in order to track the participants' movements of their fingers and hands. Jobbágy and colleagues aim at providing tests and / or measures to indicate the presence of early to moderate PD and subtle changes in its progression. Twenty-nine participants took part in their study (i.e. thirteen young healthy subjects, ten elderly healthy subjects and six subjects afflicted with PD). Three tasks were performed: tapping task, twiddling task as well as a pinching and circling task. The authors describe their analysis of raw movement data from their tracking system and highlight their chosen features (e.g. frequency, symmetry, dexterity, amplitude, etc.). Based on these parameters a score (between zero and one) is proposed in which people with PD achieve higher score-values (as in UPDRS). Their empirical results indicate that their scale does indeed separate PD patients from healthy subjects.

## 3.CONCLUSIONS

It is apparent that indication of PD motor symptoms in time series data is clearly not an unwritten page. Alone in the past decade a great number of publications with a focus on this very topic have been seen. Some of the mentioned authors have published their work on several symptoms (e.g. Cole et al. [15],[16], Salarian et al. [56],[57] and Zwartjes et al. [72]).

In recent years, accuracy of symptom indication and severity indication have reached percentages well above 90%. However it should be noted that datasets vary greatly in quality and quantity (e.g. from a few minutes to several hours or days of data). The accuracy increases and decreases with the used datasets and employed algorithms. Authors with small datasets or even synthetic datasets tend to achieve higher accuracies than those that utilize medium-sized or large datasets from real people. Another aspect of quality is the task / activity which have been performed during recording sessions (i.e. scripted vs. unscripted, constrained vs. unconstrained, etc.). Here, preference has been given to those publications that were not using standardized motor tasks to identify symptoms (and their severity). Sensitivities in the range of 90%-95% (sometimes even greater) were achieved with today's methods, but usually at the cost of a lower specificity.

Table 2 summarizes the papers that were presented in this paper. Despite the reservation of highlighting publications that enable indication of PD motor symptoms and / or assessing their severity while being mobile, a set of publications that do not fit these criteria was presented. Table 1 points to several noteworthy publications with a similar focus, but do not necessarily intend to identify cardinal symptoms or make use of body-mounted sensors. As a consequence, they do not necessarily present the state-of-the-art. Nonetheless the interested reader is encouraged to read through them.

## FUTURE WORK

Despite the fact that fairly high accuracies has already been reached, the presented results still allow for some improvements. The authors would like to employ a rather untraditional set of algorithms for indicating the presence of the mentioned PD motor symptoms in time series data (e.g. StreamKM++[2], ClusTree[33] and LogLog algorithm [22]). It is their intention to evaluate whether these approaches can perform on a similar level of accuracy or maybe even outperform the mentioned publications. A suitable framework, called MOSIS, for evaluating these approaches is actually under development (publication is pending; download available on mloss.org). Furthermore, the author's are likely to review more publications.

### ACKNOWLEDGEMENTS

This work has been motivated and partly funded by the European Commission through AAL JP Call 1 project HELP and ICT FP 7 project REMPARK (287677). The authors acknowledge the support by the Commission and the HELP and REMPARK partners for their fruitful work and contribution in the research.

Table 1. Lists several related publications. The author(s) and title of their publication are highlighted. This list is intended to supplement state-of-the-art publications (shown in Table 2) with additional noteworthy and relevant papers.

| Author(s) | Note |
| --- | --- |
| Brewer et al. [12] | Application of Modified Regression Techniques to a Quantitative Assessment for the Motor Signs of Parkinson's Disease |
| Cunningham et al. [18] | Identifying fine movement difficulties in Parkinson's disease using a computer assessment tool |
| Cunningham et al. [17] | Home-Based Monitoring and Assessment of Parkinson's Disease |
| Hamilton et al. [25] | Neural networks trained with simulation data for outcome prediction in pallidotomy for Parkinson's disease |
| Kondraske et al. [31] | Web-based evaluation of Parkinson's Disease subjects: Objective performance capacity measurements and subjective characterization profiles |
| Wang et al. [68] | A new quantitative evaluation method of Parkinson's disease based on free spiral drawing |
| Jobbágy et al. [28] | Early detection of Parkinson's disease through automatic movement evaluation |

Table 2. Summarization of state-of-the-art publications on PD symptom indication algorithms. For each symptom (T: tremor, B: bradykinesia, F: FOG, D: dyskinesia) and reference, the employed classification techniques and utilized sensors (A: accelerometer, G: gyroscope, E: EMG sensor) are highlighted. Furthermore their results are indicated. It should be kept in mind that the results among these papers are not directly comparable due to employment of different

| Symptom | Author(s) | Sensor(s) | Algorithm(s) | Result(s) | Note(s) |
|---|---|---|---|---|---|
| T | Salarian et al. [56] | G | thresholds | Sen.: 76.6% Spec.: 98.0% | Sensitivity has been averaged across pitch, roll and yaw axis. |
| T | Salarian et al. [57] | G | thresholds | Sen.: 99.5% Spec.: 94.2% | |
| T | Zwartjes et al. [72] | A, G | thresholds | Acc.: 84.7% | Average of rest tremor and kinematic across varying postures. |
| T | Rigas et al. [54] | A | HMM | Acc.: 87.0% | |
| T | Cole et al. [15] | A, E | DNN | Sen.: 93.0% Spec.: 95.0% | |
| T | Roy et al. [55] | A, E | DNN | Sen.: 91.2% Spec.: 93.4% | Average of results for arms and legs. |
| T | Niazmand et al. [46] | A | thresholds | Sen.: 80.0% Spec.: 98.5% | Average of postural and rest tremor. |
| B | Cancela et al. [14] | A | KNN, NN, SVM, etc. | Acc.: 86.5% | |
| B | Pastorino et al. [49] | A, G | SVM | Acc.: 74.4% | |
| B | Salarian et al. [56] | G | n/a | Significant for most features ($p \leq 1\%$) | |
| B | Salarian et al. [57] | G | n/a | Significant for most features ($p \leq 1\%$) | |
| B | Zwartjes et al. [72] | A, G | n/a | Significant for most features ($p \leq 2\%$) | |
| F | Djurić-Jovičić [21] | A, G | NN, thresholds | Classification error $\leq 16\%$ | |
| F | Cole et al. [16] | A, E | DNN, thresholds | Sen.: 82.9% Spec.: 97.3% | |
| F | Niazmand et al. [47] | A | thresholds | Sen.: 88.3% Spec.: 85.3% | |
| F | Bächlin et al. [11] | A, G | thresholds | Sen.: 73.1% Spec.: 81.6% | |
| F | Stamatakis et al. [62] | A | n/a | Empirical difference | |
| D | Keijsers et al. [30] | A | NN | Acc.: 96.8% | Averaged across arm, trunk and leg accuracies. |
| D | Tsipouras et al. [65] | A, G | KNN, RF, DT, etc. | Acc.: 93.7% | |
| D | Cole et al. [15] | A, E | DNN | Sen.: 91.0% Spec.: 93.0% | |
| D | Roy et al. [55] | A, E | DNN | Sen.: 90.0% Spec.: 93.4% | Averaged across patients and control subjects. |
| D | Patel et al. [51] | A | EM | Clusters well separable | |

Health Informatics- An International Journal (HIIJ) Vol.2,No.4,November 2013[37] Liu, X., Carroll, C.B., Wang, S.Y., Zajicek, J., Bain, P.G.: Quantifying druginduced dyskinesias in the arms using digitised spiral-drawing tasks. Journal of Neuroscience Methods 144(1), 47 – 52 (2005), http://www.sciencedirect.com/science/article/pii/S0165027004003723

[38] Longsta, M.G., Mahant, P.R., Stacy, M.A., Van Gemmert, A.W.A., Leis, B.C., Stelmach, G.E.: Discrete and dynamic scaling of the size of continuous graphic movements of parkinsonian patients and elderly controls. Journal of Neurology, Neurosurgery & Psychiatry 74(3), 299– 304 (2003), http://jnnp.bmj.com/content/74/3/299.abstract

[39] Marchis, C.D., Schmid, M., Conforto, S.: An optimized method for tremor detection and temporal tracking through repeated second order moment calculations on the surface emg signal. Medical Engineering & Physics 34(9), 1268 – 1277 (2012), http://www.sciencedirect.com/science/article/pii/S1350453311003444

[40] Mathers, C., Fat, D.M., Boerma, J.T., Organization, W.H.: The global burden of disease : 2004 update. World Health Organization, Geneva, Switzerland: (2008), http://www.who.int/healthinfo/global\_burden\_disease/GBD\_report\_2004update\_full.pdf

[41] Mekyska, J., Rektorova, I., Smekal, Z.: Selection of optimal parameters for automatic analysis of speech disorders in parkinson's disease. In: Telecommunications and Signal Processing (TSP), 2011 34th International Conference on. pp. 408 – 412 (aug 2011)

[42] Miralles, F., Tarongi, S., Espino, A.: Quantification of the drawing of an archimedes spiral through the analysis of its digitized picture. Journal of Neuroscience Methods 152(1 - 2), 18 – 31 (2006), http://www.sciencedirect.com/science/article/pii/S0165027005002955

[43] Myers, L.J., MacKinnon, C.D.: Quanti cation of movement regularity during internally generated and externally cued repetitive movements in patients with parkinson's disease. In: Neural Engineering, 2005. Conference Proceedings. 2nd International IEEE EMBS Conference on. pp. 281 – 284 (march 2005)

[44] Niazmand, K., Kalaras, A., Dai, H., Lueth, T.C.: Comparison of methods for tremor frequency analysis for patients with parkinson's disease. In: Biomedical Engineering and Informatics (BMEI), 2011 4th International Conference on. Vol. 2, pp. 693 – 697 (oct 2011)

[45] Niazmand, K., Tonn, K., Kalaras, A., Fietzek, U.M., Mehrkens, J.H., Lueth, T.C.: Quantitative evaluation of parkinson's disease using sensor based smart glove. In: Computer-Based Medical Systems (CBMS), 2011 24th International Symposium on. pp. 1 – 8 (june 2011)

[46] Niazmand, K., Tonn, K., Kalaras, A., Kammermeier, S., Boetzel, K., Mehrkens, J.H., Lueth, T.C.: A measurement device for motion analysis of patients with parkinson's disease using sensor based smart clothes. In: Pervasive Computing Technologies for Healthcare (PervasiveHealth), 2011 5th International Conference on. pp. 9 – 16 (may 2011)

[47] Niazmand, K., Tonn, K., Zhao, Y., Fietzek, U.M., Schroeteler, F., Ziegler, K., Ceballos-Baumann, A.O., Lueth, T.C.: Freezing of gait detection in parkinson's disease using accelerometer based smart clothes. In: Biomedical Circuits and Systems Conference (BioCAS), 2011 IEEE. pp. 201 – 204 (nov 2011)

[48] Niazmand, K., Somlai, I., Louizi, S., Lueth, T.C.: Proof of the accuracy of measuring pants to evaluate the activity of the hip and legs in everyday life. In: Lin, J.C., Nikita, K.S., Akan, O., Bellavista, P., Cao, J., Dressler, F., Ferrari, D., Gerla, M., Kobayashi, H., Palazzo, S., Sahni, S., Shen, X.S., Stan, M., Xiaohua, J., Zomaya, A., Coulson, G. (eds.) Wireless Mobile Communication and Healthcare, Lecture Notes of the Institute for Computer Sciences, Social Informatics and Telecommunications Engineering, vol. 55, pp. 235– 244. Springer Berlin Heidelberg (2011), http://dx.doi.org/10.1007/978-3-642-20865-2\_30,10.1007/978-3-642-20865-2 30

[49] Pastorino, M., Cancela, J., Arredondo, M.T., Pansera, M., Pastor-Sanz, L., Villagra, F., Pastor, M.A., Martn, J.A.: Assessment of bradykinesia in parkinson's disease patients through a multi-parametric system. In: Engineering in Medicine and Biology Society, EMBC, 2011 Annual International Conference of the IEEE. pp. 1810 – 1813 (30 2011-sept 3 2011)

[50] Patel, S., Lorincz, K., Hughes, R., Huggins, N., Growdon, J., Standaert, D., Akay, M., Dy, J., Welsh, M., Bonato, P.: Monitoring motor fluctuations in patients with parkinson's disease using wearable sensors. Information Technology in Biomedicine, IEEE Transactions on 13(6), 864 – 873 (nov 2009)

[51] Patel, S., Sherrill, D., Hughes, R., Hester, T., Huggins, N., Lie-Nemeth, T., Standaert, D., Bonato, P.: Analysis of the severity of dyskinesia in patients with parkinson's disease via wearable sensors. In: Wearable and Implantable Body Sensor Networks, 2006. BSN 2006. International Workshop on. pp. 4 pp. – 126 (april 2006)
16

Health Informatics- An International Journal (HIIJ) Vol.2,No.4,November 2013

**Authors**

**Claas Ahlrichs** is a PhD. candidate at Universitaet Bremen, where he studied computer science with a focus on artificial intelligence and wearable computing. He graduated with his thesis "Development and Evaluation of an Abstract User Interface for Performing Maintenance Scenarios with Wearable Computers" in 2011. Since 2008, Ahlrichs is involved at the Center for Computing and Communication Technologies (TZI) Technologies in the field of wearable computing. Furthermore, he has been involved in several regional and international research projects (IT-ASSIT, HELP, REMPARK) related to health care, human computer interaction and wearable computing. Currently, he works as a software developer at neusta mobile solutions GmbH in Bremen. 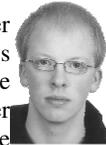

**Prof. Dr. Michael Lawo** is since 2004 at TZI (www.tzi.de) of the Universitaet Bremen, and since 2009 one of the two Managing Directors of neusta mobile solutions GmbH. He is professor for applied computer science at Universitaet Bremen, member of the steering board of Logdynamics (www.logdynamics.de) and involved in numerous projects of logistics, wearable computing and artificial intelligence. He had been the CEO of a group of SME in the IT domain since 1999 with a focus on the development and marketing of virtual reality simulators for surgeons; from 1996 to 2000 he was CEO of an IT consulting firm and from 1991 to 1995 top manager information systems with the Bremer Vulkan group. Michael Lawo was consultant before joining the nuclear research centre in Karlsruhe from 1987 to 1991 as head of the industrial robotics department. He is a 1975 graduate of structural engineering of Ruhr Universität Bochum, received his PhD from Universität Essen in 1981 and became professor in structural optimisation there in 1992. In 2000 he was appointed as professor of honour of the Harbin/China College of Administration & Management. He is author, co-author and co-publisher of eight books and more than 150 scientific papers on numerical methods and computer applications also in logistics, healthcare, optimization, IT-security, sensorial materials and wearable computing. 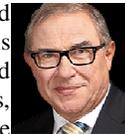